\pdfoutput=1
\documentclass[letterpaper, 10 pt, conference]{ieeeconf}

\IEEEoverridecommandlockouts
\usepackage{graphicx}
\usepackage{amsmath}
\usepackage{amssymb}
\usepackage{booktabs}
\usepackage{multirow}
\usepackage{cite}
\usepackage{algorithm}
\usepackage{algpseudocode}
\usepackage{tikz}
\usepackage{url}
\usepackage{xcolor}
\makeatletter
\def\paragraph{\@startsection{paragraph}{4}{\z@}{0ex plus 0.1ex minus 0.1ex}%
{0ex}{\normalfont\normalsize\itshape}}
\makeatother
\usetikzlibrary{arrows.meta,positioning,fit,backgrounds}

\newcommand{\ours}{ARMFUL}

\newcommand{\dataset}{WAG}

\newcommand{\nocert}{\dataset{}\textrm{-LiftOnly}}

\newlength{\pwlen}
\newcommand{\phasebrace}[2]{$\underbrace{\hspace{#1}}_{\mbox{\scriptsize #2}}$}

\title{\LARGE \bf
\ours{}: Learning Whole-Arm Grasp Configurations for Single-Arm Pickup of Large Objects
}

\author{Xinyi Yu, Yunshuang Li, Daniel Seita, Gaurav S. Sukhatme\\[2pt]
\normalsize Website: \url{https://manipulation-armful.github.io/}%
\thanks{All authors are with the Thomas Lord Department of Computer Science
at the University of Southern California, USA.}%
\thanks{Correspondence: xinyi.yu12@usc.edu}%
}

\begin{document}

\makeatletter
\g@addto@macro\@maketitle{%
\vspace{-0.4\baselineskip}%
\begin{center}%
\includegraphics[width=\textwidth]{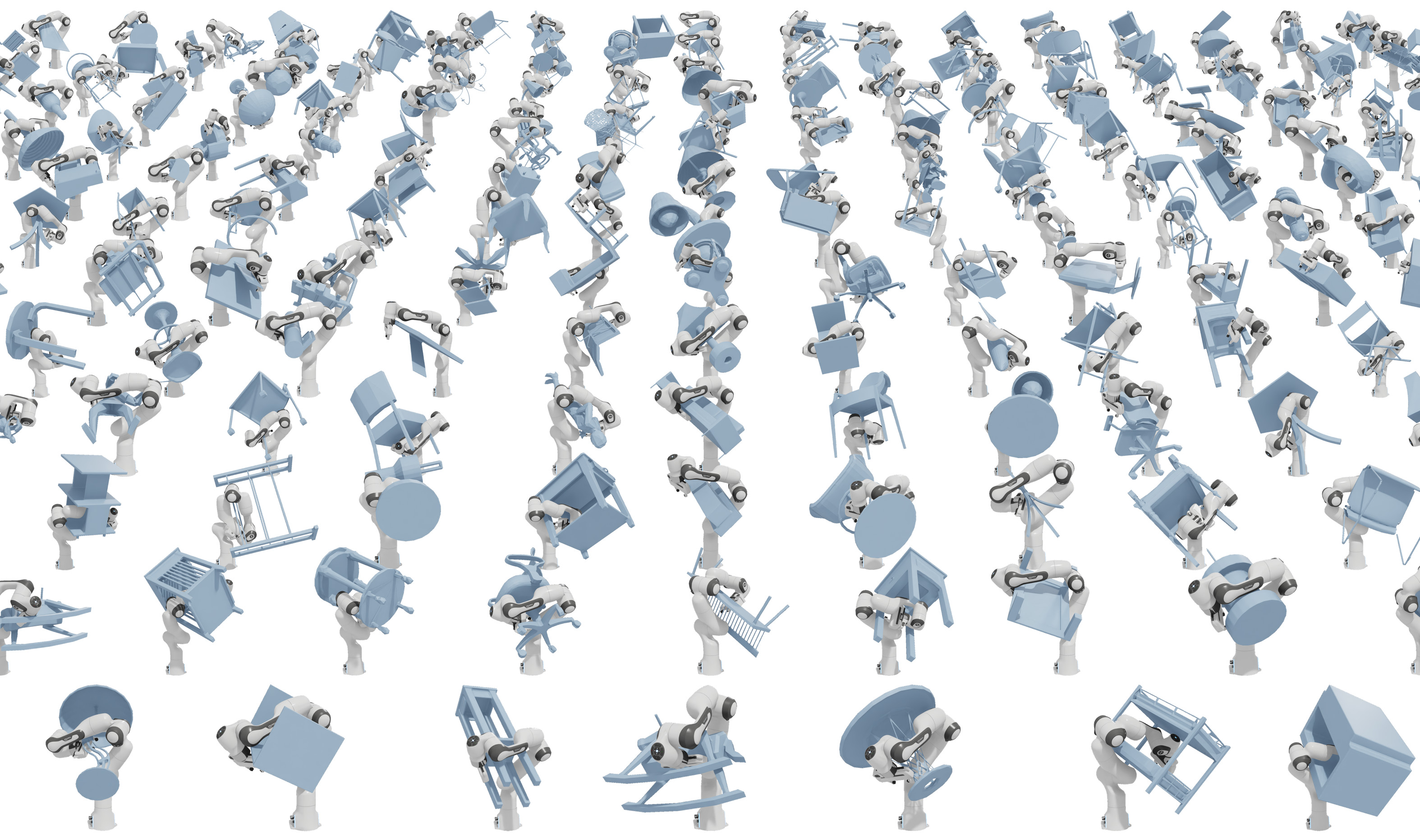}\par
\vspace{0.4\baselineskip}%
{\refstepcounter{figure}\label{fig:wag-gallery}\footnotesize\raggedright
Fig.~\thefigure.~~\textbf{Whole-arm grasps from \dataset{}.} A gallery of whole-arm grasp configurations in simulation, each shown at the hold after the
pickup execution.\par}%
\end{center}%
}
\makeatother
\maketitle
\addtocounter{figure}{-1}
\thispagestyle{plain}
\pagestyle{plain}

\begin{abstract}
Many everyday objects, such as storage boxes, baskets, and
chairs, are too large or heavy for a gripper to grasp, yet a single arm can
still pick them up by wrapping around and supporting them. Such grasps are
hard to learn: wrapping motions are difficult to teleoperate, and existing
datasets contain little whole-arm pickup data. We introduce \dataset{}
(Whole-Arm Grasp Dataset), the first large-scale dataset for single-arm pickup
of large objects, with 1.7~million joint configurations across 793
objects. We sample configurations directly in joint space, execute each through
squeeze and lift primitives in simulation, and stress-test survivors
under active shaking and external forces. To demonstrate the utility of
\dataset{} and accomplish the whole-arm pickup task, we develop \ours{}, a whole-arm pickup pipeline that pairs a diffusion model trained on \dataset{}, generating whole-arm grasps from the object's point cloud, with a motion planner that reaches them collision-free before the same squeeze-and-lift execution. In simulation, \ours{} reaches 85.4\% lift success
on training and 69.9\% on held-out objects, and it lifts the
object in 70 of 100 trials on a physical Franka arm across 10 IKEA objects.
\end{abstract}

\section{Introduction}
\label{sec:intro}

Robotic grasping has made substantial progress with a range of
end-effectors, such as parallel-jaw
grippers~\cite{fang2020graspnet}, dexterous
hands~\cite{wang2023dexgraspnetlargescaleroboticdexterous}, and suction
grippers~\cite{cao2021suctionnet}. However, end-effector-centric grasping
remains
fundamentally limited when objects are too large to be securely grasped
or their weight exceeds the end-effector's payload limit, e.g., chairs, tables, and large storage boxes. Picking up and transporting them requires
using the entire arm,
motivating \emph{whole-arm manipulation} to overcome the geometric,
kinematic, and force limitations of end-effector-only manipulation.

Unlike conventional grasping, a whole-arm grasp cannot be specified by
a single end-effector pose. Its success depends on the full joint
configuration and contacts established along multiple links, making the
interaction more sensitive than conventional small-object grasping to
the object's geometry and physical properties, such as mass distribution
and surface friction. This makes it difficult to design grasping
strategies that generalize across objects. Learning offers a promising
alternative but requires whole-arm grasp data. Existing large-scale
manipulation datasets primarily capture end-effector interactions
(Section~\ref{sec:related}). Large-scale simulated datasets with
automatically generated labels have supported learning-based gripper and
dexterous-hand grasping~\cite{wang2023dexgraspnetlargescaleroboticdexterous,zhang2024dexgraspnet20learninggenerative,ye2025dex1blearning1bdemonstrations,karim2025dg16m,fang2020graspnet,cao2021suctionnet,mahler2018dex},
but comparable data for whole-arm grasping remain unavailable.

\begin{figure*}[t]
\centering
\includegraphics[width=\textwidth]{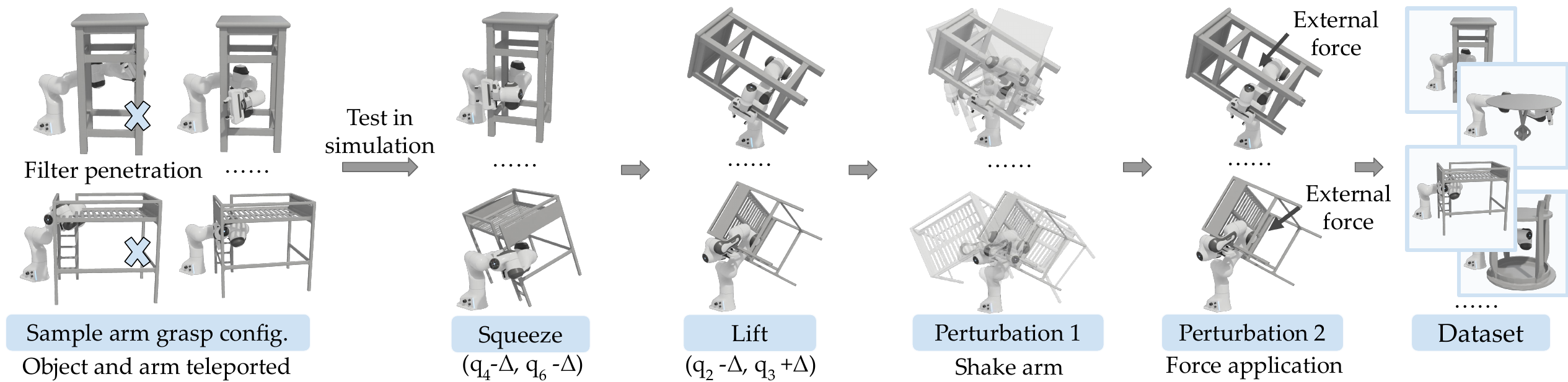}
\caption{\textbf{Construction of \dataset{}}. We sample candidate whole-arm grasp
configurations, discard penetrating candidates, and test the rest
directly in a GPU-parallelized simulator. The arm is teleported to the
candidate grasp configuration and executes the squeeze and
lift primitives. Configurations that hold the
object throughout the lift and maintain the grasp with base-joint shaking and external force applied to the object are added to \dataset{}.}
\label{fig:wag}
\end{figure*}

To fill this gap, we first introduce \dataset{} (the \textbf{W}hole-\textbf{A}rm
\textbf{G}rasp Dataset), a large-scale dataset generated and stress-tested
entirely in simulation (Fig.~\ref{fig:wag}), together with \ours{}, a
learning-and-planning pipeline for single-arm pickup of large objects
(Fig.~\ref{fig:armful}). To build \dataset{}, we sample whole-arm grasp
configurations in joint space and test the candidates in a GPU-parallel
simulation using two scripted primitives: \emph{squeeze} and \emph{lift}.
Only joint angles that survive arm shaking and external forces are
retained.
To demonstrate the usefulness of this dataset, we develop \ours{}, which uses a diffusion model trained on \dataset{} to learn whole-arm grasp configurations from an object's point cloud.
It then plans a
collision-free approach to a learned configuration and executes the
same squeeze-and-lift sequence.
In simulation, \ours{} outperforms heuristic and retrieval
baselines. We also validate the pipeline on 10 IKEA objects in the real world.
Our key contributions are as follows:
\begin{itemize}[\setlength{\labelindent}{0pt}]
\item We construct \dataset{}, the first large-scale whole-arm grasp dataset for single-arm pickup of large objects. It contains 793 diverse objects and approximately 1.7 million grasp configurations.

\item We introduce a perturbation-based stress test that evaluates each grasp and jittered copies of it under base-joint shaking and external forces, retaining only robust configurations and helping increase the dataset quality.

\item We demonstrate the utility of \dataset{} and accomplish the
full whole-arm pickup task through \ours{}, a
whole-arm grasping pipeline trained on the dataset. \ours{} combines
a diffusion model over grasp configurations with collision-free
approach planning. It reaches 85.4\% and 69.9\% lift success on training
and held-out objects in simulation, outperforming heuristic and
retrieval baselines, and lifts the object in 70 of 100 trials on a
physical robot.
\end{itemize}

\section{Related Work}
\label{sec:related}

\paragraph{Large Object Manipulation}
Prior work has explored large-object manipulation with several robot
embodiments and technical approaches. Humanoid and compliant robots
learn whole-body box pickup, carrying, and embracing through
sim-to-real or example-guided reinforcement
learning~\cite{dao2024simtoreal,zheng2025embracing,barreiros2025learning}.
Loco-manipulation systems such as RobotMover exploit a mobile base to
reposition contacts while moving furniture-scale
objects~\cite{li2025robotmoverlearninglargeobjects}. Bimanual
systems distribute the load across two arms via closed-chain motion
planning or planning-guided diffusion
policies~\cite{zhou2017closedchain,li2025planningguideddiffusionpolicylearning}.
On the planning side, physics-informed eikonal caging offers a
differentiable enclosure objective for whole-arm
manipulation~\cite{zhang2026physics}, and force-controlled assistance
with large loads also arises in
caregiving~\cite{kowalski2023rulebased}. These systems rely on a
humanoid or mobile base, multiple arms, or task-specific support
motions, or target secure enclosure during manipulation rather than
pickup. We instead study large-object pickup with one arm
and use contacts along its links to form the grasp.

\paragraph{Manipulation Datasets}
Large-scale datasets have been a key driver of recent progress toward
scalable and general robotic manipulation
policies~\cite{FMs_robotics_2023_survey_1,FMs_robotics_2023_survey_2,real_world_robot_FMs_2024}.
Real-world collections such as Open
X-Embodiment~\cite{embodimentcollaboration2025openxembodimentroboticlearning}, ABC~\cite{allshire2026scalable},
and DROID~\cite{khazatsky2025droidlargescaleinthewildrobot} aggregate
large numbers of teleoperated trajectories across robots and scenes,
and underpin recent Vision-Language-Action
systems~\cite{geminiroboticsteam2025geminirobotics15pushing,
intelligence2025pi06vlalearnsexperience,
lee2025molmoactactionreasoningmodels}.
Simulation-generated datasets
complement them with dense, automatically labeled supervision at far
lower collection cost, covering suction
grasping~\cite{mahler2018dex}, single-object and cluttered-scene
dexterous
grasping~\cite{wang2023dexgraspnetlargescaleroboticdexterous,zhang2024dexgraspnet20learninggenerative},
dexterous grasping and
articulation~\cite{ye2025dex1blearning1bdemonstrations}, and dual-arm
grasping~\cite{karim2025dg16m}.
However, these resources remain almost
exclusively end-effector-centric: contact is confined to the
parallel-jaw gripper, suction gripper, or dexterous hand. To the best of
our knowledge,
\dataset{} is the first large-scale dataset to target single-arm pickup of large
objects with whole-arm grasp configurations.

\paragraph{Whole-Arm Manipulation}
Whole-arm manipulation (WAM) has been studied both as an alternative
to gripper-centric manipulation for objects that exceed gripper limits
in size or weight, and for tasks that inherently require contact along
the arm, such as physically assisting people. Prior work has
investigated WAM through multi-contact
modeling~\cite{song2000dynamic}, caging-based
planning~\cite{zhang2026physics}, learning from human demonstrations~\cite{huang2026wacl},
imitation~\cite{fang2024airexo} or reinforcement
learning~\cite{barreiros2025learning,zheng2025embracing,mcgartoll2026wholebody},
contact-centric predictive control with distributed tactile
sensing~\cite{madan2026tactic}, contact-aware
planning~\cite{li2025robotmoverlearninglargeobjects,
li2025planningguideddiffusionpolicylearning}, humanoid
loco-manipulation~\cite{xie2026grail,cao2026weave}, and hardware designs
tailored for whole-arm
interaction~\cite{xu2025robopanoptesallseeingrobotwholebody}.
Collectively, these works establish WAM as an effective paradigm for
large-object handling, but typically require explicit modeling or
dedicated training of whole-arm behaviors as standalone skills. In
contrast, \ours{} learns whole-arm grasp configurations from \dataset{},
generalizing across diverse
large objects without object-specific modeling or per-skill training.

\section{Problem Statement and Assumptions}
\label{sec:problem}

We study the pickup of large objects from a flat support surface using
a single fixed-base 7-DoF manipulator (the Franka Panda) equipped with
its stock parallel-jaw gripper.
We assume that one object $O$ rests on the
surface in a stable configuration $S_{\mathrm{obj}} \in SE(3)$ and that
the object lies within the arm's reachable workspace.
We use $P \in \mathbb{R}^{N \times 3}$ to denote the object's point cloud, a
fixed-size set of $N = 512$ points with $(x, y, z)$ coordinates, obtained from depth cameras and expressed in the robot's base frame, so
that $P$ jointly captures the object's shape, position, and orientation.
Let $\mathbf{q} = (q_1, \dots, q_7) \in \mathbb{R}^7$ denote the vector of the robot arm's seven joint angles, and let $\mathbf{q}_0$ denote its given initial value.
A \emph{whole-arm grasp configuration} $\mathbf{q}_{\mathrm{grasp}}$ is a static arm state in
which the arm encloses the object without penetrating it, such that
closing the wrap establishes the contacts that hold the object.
We use \emph{grasp} in a broad sense. The arm holds the object through
contacts along several links, which may wrap around it, hook under a
frame member, or carry part of its weight on the forearm, and we do not
require force closure. Whether a configuration holds the object is
judged by its outcome. A \emph{trial} is a complete pickup attempt consisting of an arm
trajectory from the initial configuration $\mathbf{q}_0$ to a grasp
configuration $\mathbf{q}_{\mathrm{grasp}}$, followed by a lifting
motion to a raised configuration $\mathbf{q}_{\mathrm{raised}}$ as
the wrap closes and the object leaves the support surface.
A trial is considered successful if the object is lifted off the
support surface and held there stably
(Section~\ref{sec:setup-metrics}).
Given the object observation
$P$ and the initial configuration $\mathbf{q}_0$, our objective is to
generate an arm trajectory through $\mathbf{q}_{\mathrm{grasp}}$ and
$\mathbf{q}_{\mathrm{raised}}$ that realizes a successful trial.

\begin{figure*}[t]
\centering
\includegraphics[width=\textwidth]{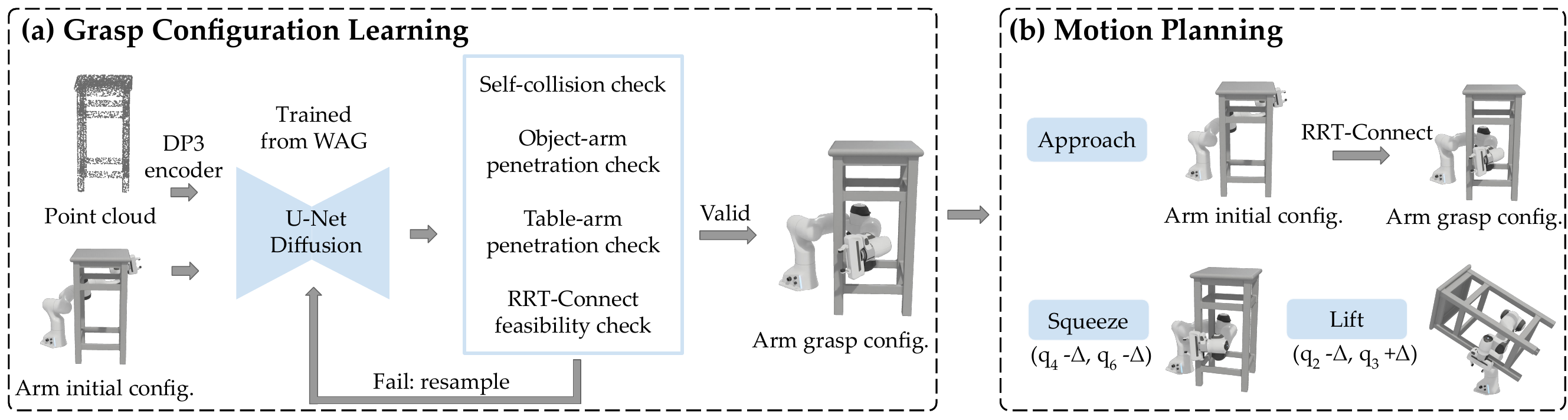}
\caption{\textbf{Overview of \ours{}}. \textbf{(a) Grasp configuration learning.} A diffusion model conditioned
on the object's point cloud, encoded with the DP3 encoder, and on the arm's initial
configuration generates whole-arm grasp configurations (the arm's seven joint angles), which are filtered
for self-collision, object and table penetration, and planner feasibility.
\textbf{(b) Motion planning.} The arm follows an RRT-Connect path to the
selected arm grasp configuration, and the same squeeze-and-lift execution completes the pickup.}
\label{fig:armful}
\end{figure*}
\section{Method}
\label{sec:method}

We first describe how \dataset{} is generated
(Section~\ref{sec:method-data}; Fig.~\ref{fig:wag}), then the \ours{}
pipeline that uses it (Fig.~\ref{fig:armful}). \ours{} has two stages.
A conditional diffusion model learns to generate grasp configurations
from the object's point cloud (Section~\ref{sec:method-learning};
Fig.~\ref{fig:armful}a), and a collision-aware motion planner reaches
a generated configuration at deployment, from which the squeeze and
lift primitives pick up the object
(Section~\ref{sec:method-planning}; Fig.~\ref{fig:armful}b).

\subsection{Dataset Generation}
\label{sec:method-data}

We build \dataset{} based on the object meshes of
DG16M~\cite{karim2025dg16m}, a dual-arm grasping dataset whose objects
are chosen to be too large for a single parallel-jaw
gripper. From these meshes we keep the objects whose size fits within
the arm's workspace and that rest stably on the tabletop in
simulation.

For each stable object placement (one of the object's stable resting
orientations, a yaw from a uniform grid, and a random tabletop position), we sample arm configurations
directly in joint space.
Forward kinematics places the arm's collision meshes, and we discard any
configuration that penetrates the object or the tabletop, or whose
forearm and wrist links are not within wrapping proximity of the object
surface. Those that survive are taken as candidate grasp configurations.

Each joint configuration candidate is then tested in physics simulation. We initialize the
arm directly to the candidate configuration, with the object at its
settled pose, so that tens of thousands of candidates are evaluated
simultaneously in parallel environments. This skips the approach from
$\mathbf{q}_0$ to $\mathbf{q}_{\mathrm{grasp}}$, as \dataset{} is a dataset of grasp
configurations. From this candidate grasp configuration
$\mathbf{q}_{\mathrm{grasp}}$, the arm is executed through scripted
\emph{squeeze} and \emph{lift} primitives.
The \emph{squeeze}
primitive closes the wrap around the object by commanding an
inward offset on the wrist joint $q_6$ and the elbow joint
$q_4$, and the \emph{lift} primitive then raises the object off the
tabletop with offsets on the shoulder joint $q_2$ and the
upper-arm joint $q_3$, signed so the arm lifts from whichever side of
the object it started on. These are PD targets instead of
trajectories the arm follows, so once the links contact the object the
arm stalls short of the commanded offset, and the residual tracking
error produces the squeezing force.
A candidate configuration
passes the nominal test if the resulting trial is successful, as
defined in Section~\ref{sec:problem}. Notably, the object is not held
rigidly. Since raising it requires rotating the arm, the object rolls
inside the wrap and settles into a different attitude from
the one at which it was grasped. Passing the nominal test therefore
already requires the wrap to hold the object through substantial
relative motion.

Each candidate configuration that passes the nominal test is then
\emph{stress-tested} over five trials, the nominal configuration and four
jittered copies offset by up to $\pm0.05$\,rad per joint. In every trial, the lifted grasp is shaken back and forth about the base joint $q_1$ for
five cycles of $\pm0.3$\,rad at $1.1$\,Hz, near its velocity limit, and a
$30$\,N external force is applied to the object for
$0.5$\,s, directed out of the wrap.
The direction is fixed in the frames of the two links that
form the wrap rather than in the world frame, so the disturbance is
the same relative to the grasp for every object and pose. Concretely,
it bisects the forearm link's local $+y$ axis and the upper-arm
link's local $-z$ axis.
A candidate grasp configuration $\mathbf{q}_{\mathrm{grasp}}$ is collected into \dataset{}
only if all five trials retain the object through the lift, the shake,
and the force. These steps make \dataset{} a
high-quality grasping dataset. The shake and the external force keep only
reliable wraps, rejecting those that hold under the gentle,
quasi-static motion of the nominal lift alone or that survive by
exploiting transient artifacts of the simulator's contact resolution.
Repeating the test over jittered copies keeps only robust
configurations, rejecting those that succeed at the exact sampled
joint values alone.

To add more stress-tested configurations per placement, we then run
three more rounds that resample around the configurations that have passed the
stress test so far, adding uniform noise to their joints, and test the new candidates
in the same way.
The resulting dataset contains
around 1.7~million stress-tested grasp configurations across
793 objects.
The object pose varies across configurations, of which each object has
around 2.2k on average. Each places the object in one of its stable resting
orientations, at a yaw from a uniform grid and a position sampled over the workspace.
All data generation took about three and a half days on a single NVIDIA GeForce RTX 5090 GPU (32GB).

\subsection{Grasp Configuration Learning}
\label{sec:method-learning}

Based on \dataset{}, we train a diffusion model to generate grasp configurations. The distribution of valid configurations is inherently multimodal, as a single placement typically admits multiple distinct wraps on either side of the object.
Because diffusion models are well suited to modeling multimodal distributions ~\cite{chi2023diffusionpolicy}, we adopt a DDPM~\cite{ho2020denoisingdiffusionprobabilisticmodels} over the 7-DoF joint configuration.

The model is conditioned on the object observation and on the arm's
initial configuration $\mathbf{q}_0$.
The point cloud $P$, expressed in the robot frame, is encoded
with the lightweight point cloud encoder of 3D Diffusion Policy
(DP3)~\cite{ze2024dp3}, in which a shallow per-point MLP followed by
max-pooling and a linear projection produces a compact embedding of
the cloud. Because the points are kept in the robot frame rather than
centered per object, the embedding captures the object's
shape, position, and orientation.
We encode $\mathbf{q}_0$ by its binary grasp side, according to whether the
elbow lies to the left or right of the arm base in $\mathbf{q}_0$. A learned embedding of this side is concatenated with the
point-cloud embedding.
This side strongly constrains which wraps the arm
can form, so conditioning on it lets the model target the reachable
ones. We use a convolutional U-Net as the denoiser, conditioned on this
embedding and trained with the standard denoising objective.
Training on the training objects of \dataset{} takes about 10~hours
(3{,}000 epochs) on the same RTX 5090 GPU.

\subsection{Motion Planning}
\label{sec:method-planning}

At deployment on an object, the model first generates an arm grasp configuration $\mathbf{q}_{\mathrm{grasp}}$
conditioned on the object point cloud $P$ and the arm's initial configuration $\mathbf{q}_0$,
and each candidate is filtered before execution. Since a generated
configuration can slightly penetrate the object, a candidate that comes
within a few millimeters of the object or the tabletop is first pushed
out by small joint-space steps that reduce its penetration. We run self-collision, arm-object
penetration and arm-table penetration checks at
$\mathbf{q}_{\mathrm{grasp}}$, rejecting candidates that fail any of them.
A collision-free approach path from $\mathbf{q}_0$ to $\mathbf{q}_{\mathrm{grasp}}$ is then planned with
RRT-Connect~\cite{kuffner2000rrtconnect}. Both the checks and the
planner represent the object by the fused point cloud on hardware and
by its collision geometry in simulation.
If a candidate is rejected, or RRT-Connect finds no path to it, we
draw a new sample from the model; after 10 unsuccessful draws the
trial is counted as a failure. Once a path is found, the arm follows
it to $\mathbf{q}_{\mathrm{grasp}}$ and executes the squeeze and lift
primitives of Section~\ref{sec:method-data}.

\section{Experimental Setup}
\label{sec:setup}

Our experiments are designed to show both simulation and
real-world performance.
\textit{Simulation Performance:} how well does a whole-arm grasp
configuration learned entirely in simulation, executed through our
\emph{approach$\,\to\,$squeeze$\,\to\,$lift} pipeline, perform across
diverse objects, how does it compare to heuristic and retrieval
baselines, and does the stress test
applied during dataset generation improve the learned grasp
configurations?
\textit{Real-World Performance:} how well does the pipeline
grasp real objects on a physical robot, and what do the same objects
at the same placements do in simulation?
The remainder of this section describes the simulation and dataset,
the evaluation object sets, the real-world setup, the baselines, and the
metrics. Section~\ref{sec:results} reports the corresponding results.

\begin{figure*}[t]
\centering
\footnotesize
\setlength{\tabcolsep}{1pt}
\begin{tabular}{cccccccccc}
\includegraphics[width=0.096\linewidth]{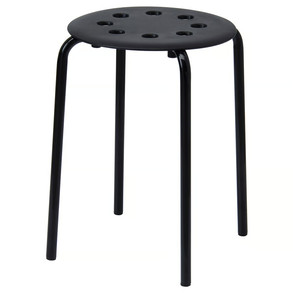} &
\includegraphics[width=0.096\linewidth]{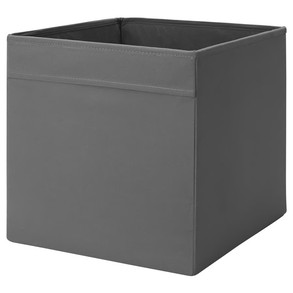} &
\includegraphics[width=0.096\linewidth]{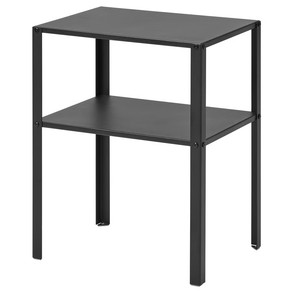} &
\includegraphics[width=0.096\linewidth]{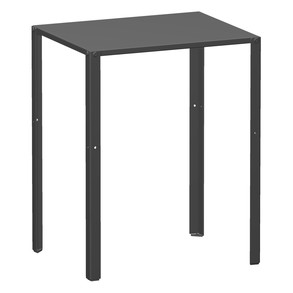} &
\includegraphics[width=0.096\linewidth]{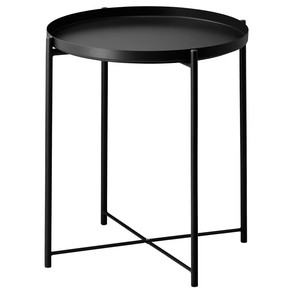} &
\includegraphics[width=0.096\linewidth]{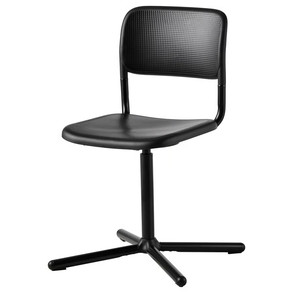} &
\includegraphics[width=0.096\linewidth]{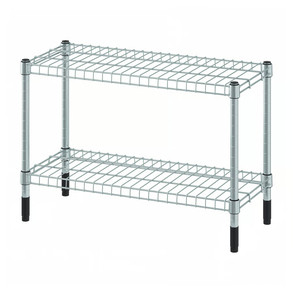} &
\includegraphics[width=0.096\linewidth]{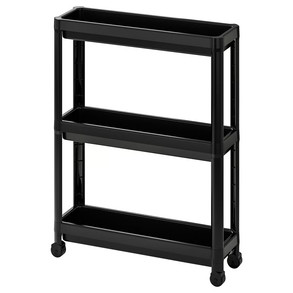} &
\includegraphics[width=0.096\linewidth]{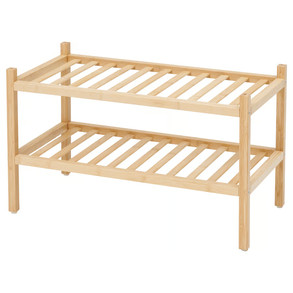} &
\includegraphics[width=0.096\linewidth]{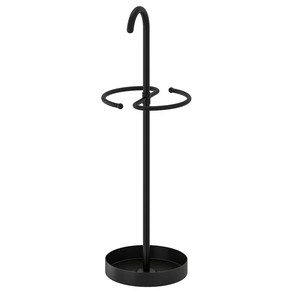} \\
Object 1 & Object 2 & Object 3 & Object 4 & Object 5 & Object 6 & Object 7 & Object 8 & Object 9 & Object 10 \\
\end{tabular}
\caption{The 10 real-world evaluation objects, all commercially
available IKEA items, except Object~4, which is
Object~3 with its middle shelf removed.}
\label{fig:real-objects-photos}
\end{figure*}

\subsection{Simulation Environment and Dataset}
\label{sec:setup-sim}

We generate training data and evaluate \ours{} in Isaac
Lab~\cite{mittal2025isaaclab}, NVIDIA's robot-learning framework built on
Isaac Sim~\cite{nvidia2025isaacsim}, using the GPU-parallel PhysX
backend~\cite{makoviychuk2021isaacgym}.
While the data
generation and grasping pipeline is not tied to a particular robot and
can be applied to other arms, we take the Franka Panda 7-DoF arm with
its stock parallel-jaw gripper as an example to detail our data
generation and pipeline evaluation.
Following Section~\ref{sec:method-data}, we build \dataset{} from
DG16M~\cite{karim2025dg16m}, obtaining around 1.7~million stress-tested
grasp configurations across 793 objects.

\subsection{Evaluation Object Sets}
\label{sec:setup-real}

We evaluate on three groups of objects; the latter two are unseen
during training:
\begin{itemize}[\setlength{\labelindent}{0pt}]
\item \textbf{Training object set}: 140 objects drawn at random from those used to train
\ours{}, evaluated in simulation only.
\item \textbf{Held-out object set}: the 140 of \dataset{}'s 793 source objects
held out from training, evaluated in simulation only.
\item \textbf{IKEA object set}: 10 objects from the IKEA catalog that are not
included in \dataset{} and form our real-world evaluation set. Each is
a commercially available item whose exact 3D model is publicly
downloadable from the IKEA website, allowing the same object to be
evaluated in simulation and on the physical robot.
Figure~\ref{fig:real-objects-photos} shows the full set.
\end{itemize}

\subsection{Real-World Setup}
\label{sec:setup-real-cell}

The physical setup is a fixed-base Franka Panda on a table, observed by
five depth cameras mounted around it, three Intel RealSense and two ZED
units. Their extrinsics come from hand-eye calibration, so every cloud
is expressed in the robot's base frame and the five views fuse into a
single observation. On hardware the partial point cloud these cameras
return is what the model receives as $P$, in place of the complete
surface available in simulation.

\subsection{Baselines}
\label{sec:setup-baselines}

We compare the simulation performance of \ours{} against two
alternative methods for producing a whole-arm grasp configuration,
evaluated in simulation on the training and held-out object sets of
Section~\ref{sec:setup-real}. Section~\ref{sec:results-q1} reports the
comparison.

\begin{itemize}[\setlength{\labelindent}{0pt}]
\item \textbf{Middle-height wrap}: a hand-designed heuristic that always
wraps the object at half its height, without adapting the wrap to the
rest of its shape. The wrap is executed in three stages. The base joint
$q_1$ first sweeps the arm inward around
the object's horizontal cross-section, slowly closing the gap and
stopping just short of contact; the elbow joint $q_4$ then closes
the remaining gap; finally the wrist joint $q_6$ turns the hand
inward to complete the wrap.
\item \textbf{Nearest-neighbor (NN) retrieval}: given a query object, retrieve its nearest
neighbor in the training set by the Basis Point Set (BPS)~\cite{prokudin2019BPS} of its point cloud and replay a
configuration randomly selected among that neighbor's recorded grasp
configurations. BPS encodes a point cloud as a fixed-length vector of
distances from a fixed set of basis points to their nearest points in
the cloud, computed here in the robot frame as in
Section~\ref{sec:method-learning}, so the descriptor captures the
object's position and orientation as well as its shape and the
database is indexed by placement rather than by object.
Since the training objects are already in this database, we
evaluate retrieval on the held-out set only.
\end{itemize}

\begin{table}[t]
\caption{Simulation results. Success rate (\%), mean $\pm$ standard
deviation over five evaluation seeds.}
\label{tab:sim}
\centering
\footnotesize
\setlength{\tabcolsep}{2.7pt}
\begin{tabular}{@{}lcccc@{}}
\toprule
& \multicolumn{2}{c}{Training object}
& \multicolumn{2}{c}{Held-out object} \\
\cmidrule(lr){2-3} \cmidrule(lr){4-5}
Method & Lift & Perturb. & Lift & Perturb. \\
\midrule
Middle-height wrap
& 54.63$\pm$1.03 & 46.43$\pm$1.21
& 54.29$\pm$1.07 & 45.46$\pm$1.55 \\
NN retrieval
& --\textsuperscript{\(\dagger\)} & --\textsuperscript{\(\dagger\)}
& 58.00$\pm$1.17 & 52.89$\pm$0.67 \\
\midrule
\ours{} (LiftOnly)
& 81.49$\pm$1.27 & 76.71$\pm$2.24
& 67.40$\pm$1.39 & 61.60$\pm$1.71 \\
\ours{}
& \textbf{85.40$\pm$0.85} & \textbf{81.80$\pm$1.47}
& \textbf{69.86$\pm$1.60} & \textbf{63.83$\pm$1.93} \\
\bottomrule
\end{tabular}

\vspace{2pt}
\begin{minipage}{\columnwidth}
\scriptsize
\textsuperscript{\(\dagger\)}Not evaluated on the training objects,
which are already in the retrieval database.
\end{minipage}
\end{table}

\begin{table*}[t]
\caption{The 10 real-world objects on hardware and in simulation.}
\label{tab:q3}
\centering
\footnotesize
\setlength{\tabcolsep}{4pt}
\begin{tabular}{lccccccccccc}
\toprule
& \multicolumn{10}{c}{Object} & \\
\cmidrule(lr){2-11}
& 1 & 2 & 3 & 4 & 5 & 6 & 7 & 8 & 9 & 10 & All \\
\midrule
Real & 10/10 & 9/10 & 6/10 & 6/10 & 7/10 & 5/10 & 9/10 & 3/10 & 7/10 & 8/10 & 70/100 \\
\midrule
Sim, replayed real trajectory & 10/10 & 4/10 & 10/10 & 9/10 & 5/10 & 9/10 & 9/10 & 3/10 & 8/10 & 9/10 & 76/100 \\
Sim, \ours{} (full pipeline) & 6/10 & 8/10 & 7/10 & 10/10 & 9/10 & 7/10 & 9/10 & 4/10 & 10/10 & 9/10 & 79/100 \\
\bottomrule
\end{tabular}
\end{table*}

\begin{figure*}[t]
\centering
\footnotesize
\setlength{\tabcolsep}{0.5pt}
\newcommand{\cw}{0.098\textwidth}
\begin{tabular}{cccccccccc}
  \includegraphics[width=\cw]{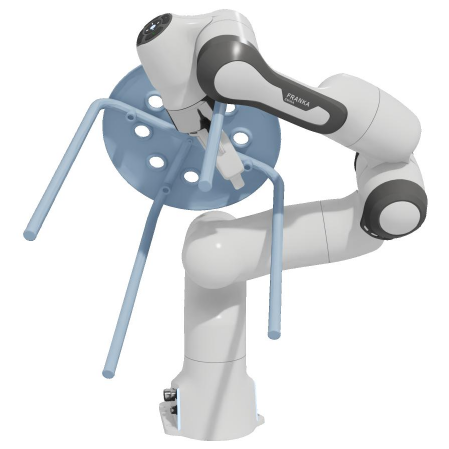} &
  \includegraphics[width=\cw]{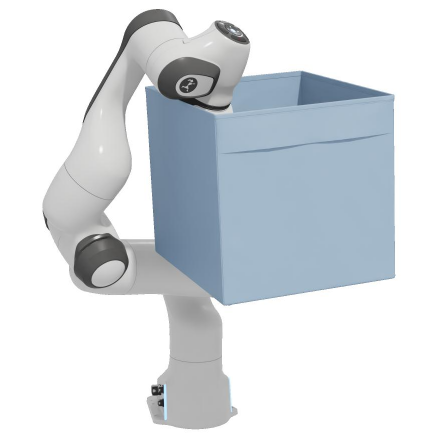} &
  \includegraphics[width=\cw]{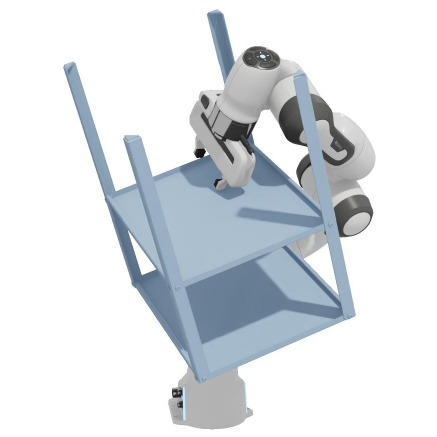} &
  \includegraphics[width=\cw]{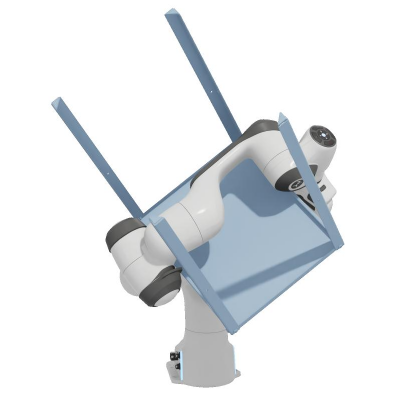} &
  \includegraphics[width=\cw]{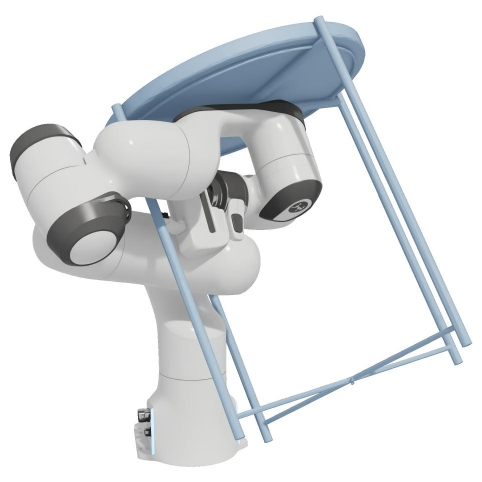} &
  \includegraphics[width=\cw]{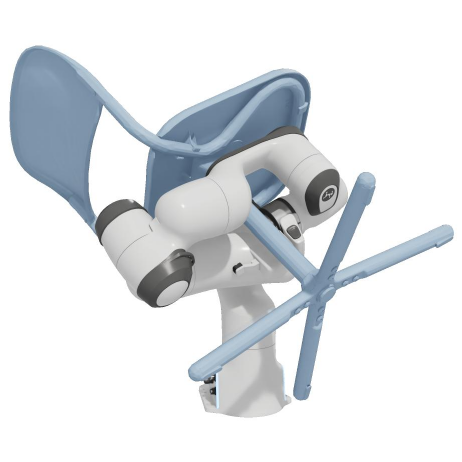} &
  \includegraphics[width=\cw]{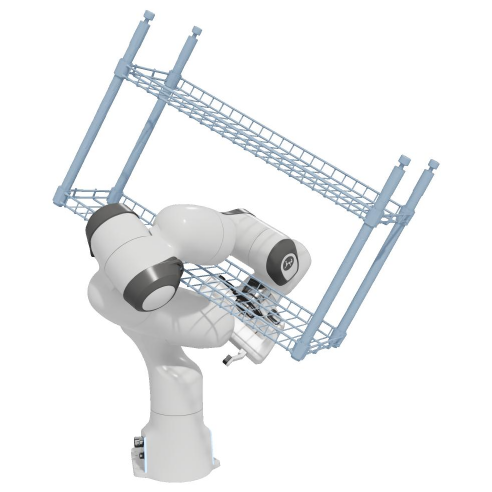} &
  \includegraphics[width=\cw]{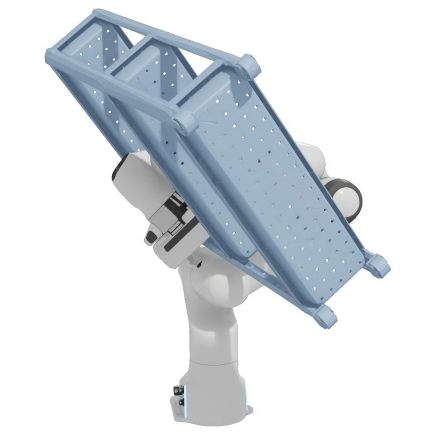} &
  \includegraphics[width=\cw]{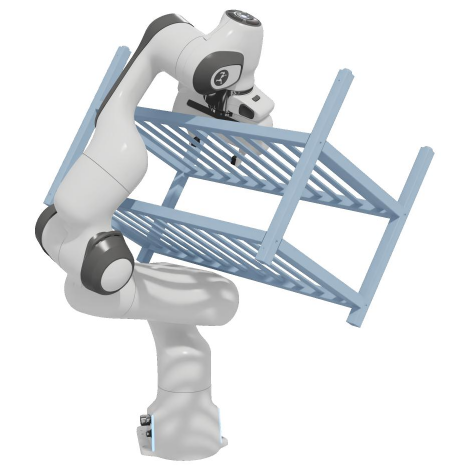} &
  \includegraphics[width=\cw]{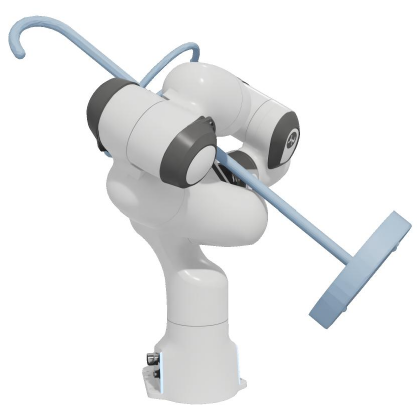} \\[1pt]
  \includegraphics[width=\cw]{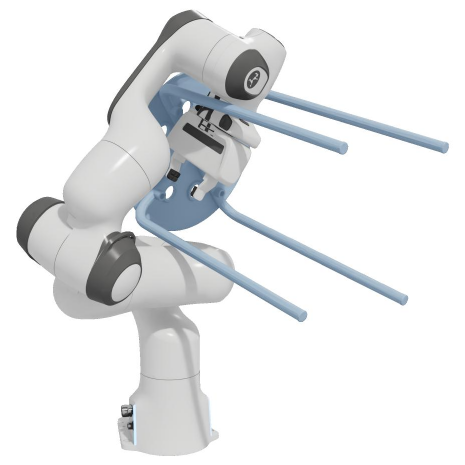} &
  \includegraphics[width=\cw]{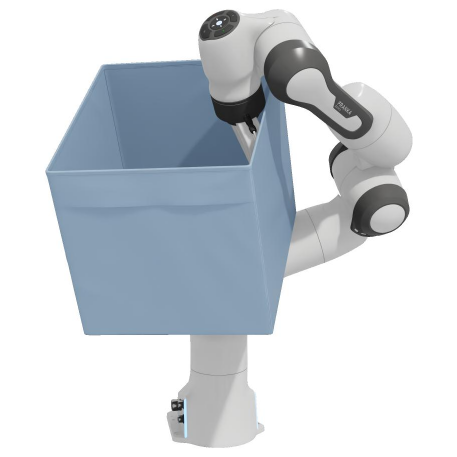} &
  \includegraphics[width=\cw]{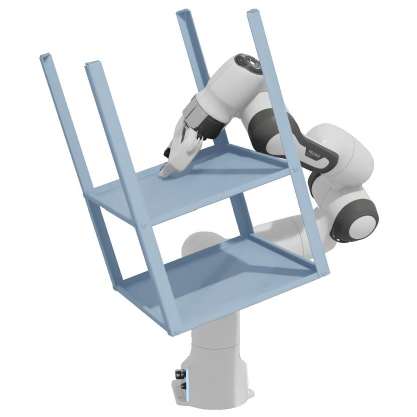} &
  \includegraphics[width=\cw]{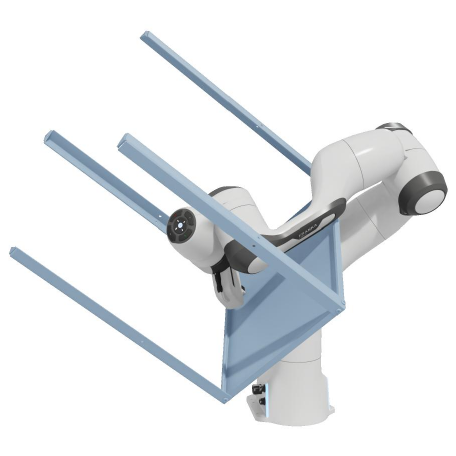} &
  \includegraphics[width=\cw]{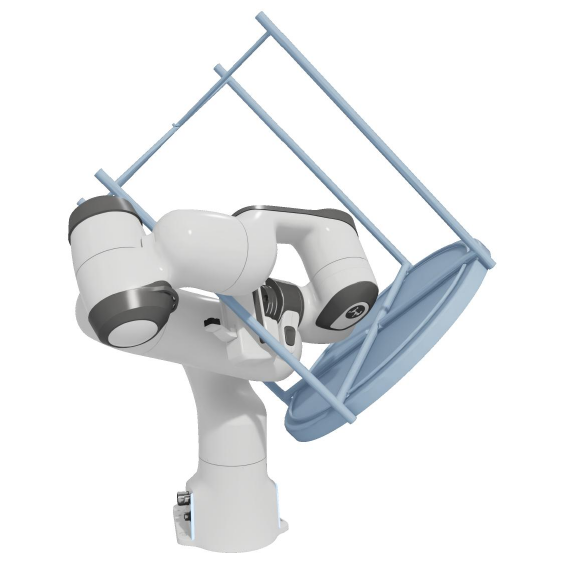} &
  \includegraphics[width=\cw]{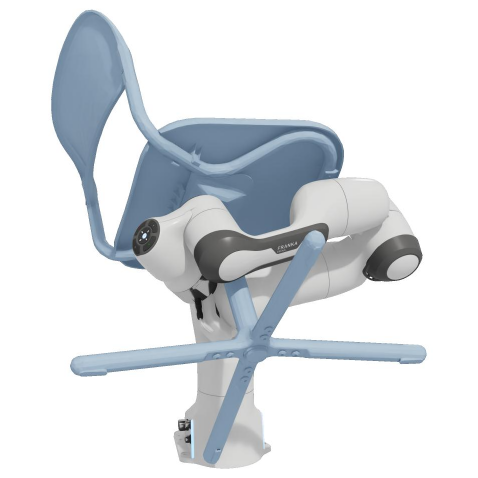} &
  \includegraphics[width=\cw]{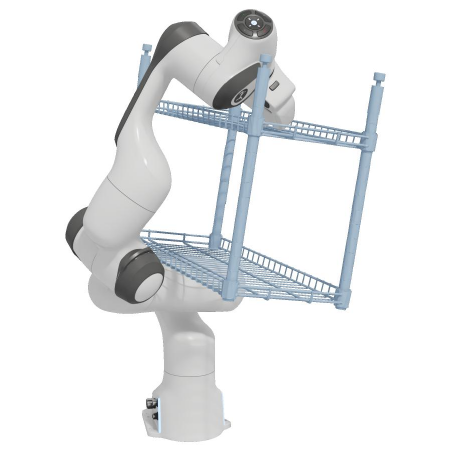} &
  \includegraphics[width=\cw]{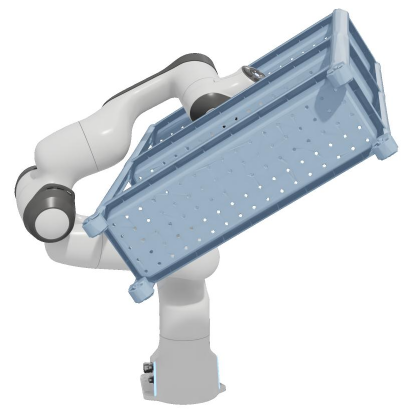} &
  \includegraphics[width=\cw]{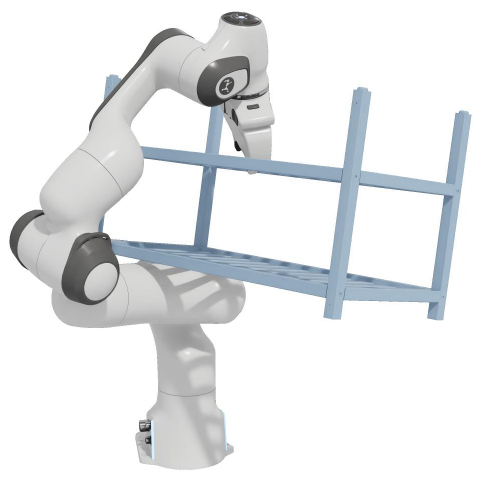} &
  \includegraphics[width=\cw]{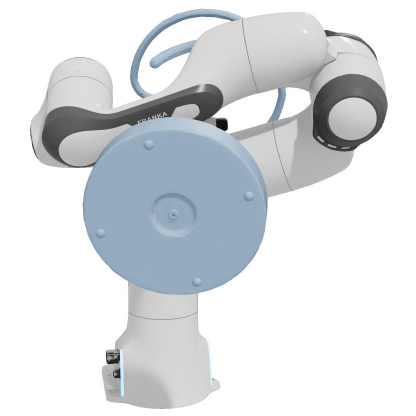} \\[3pt]
  \includegraphics[width=\cw]{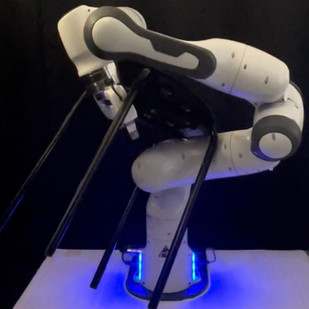} &
  \includegraphics[width=\cw]{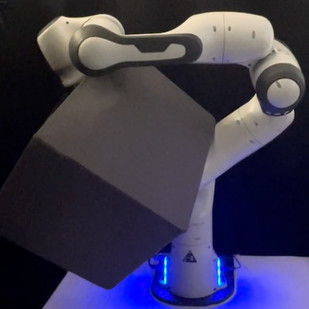} &
  \includegraphics[width=\cw]{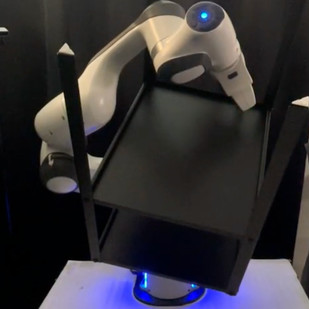} &
  \includegraphics[width=\cw]{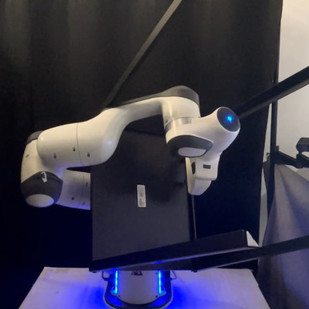} &
  \includegraphics[width=\cw]{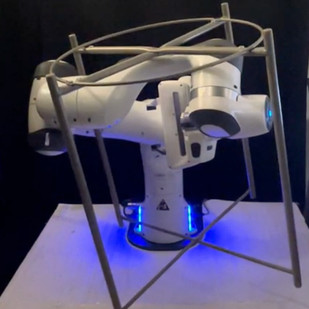} &
  \includegraphics[width=\cw]{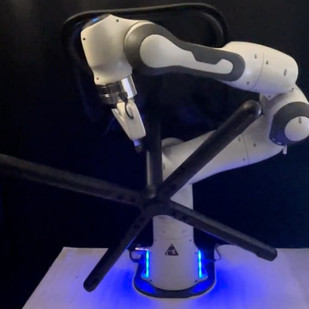} &
  \includegraphics[width=\cw]{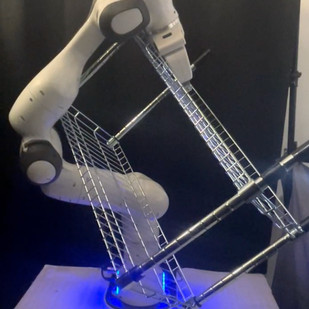} &
  \includegraphics[width=\cw]{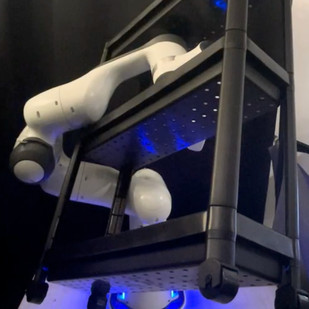} &
  \includegraphics[width=\cw]{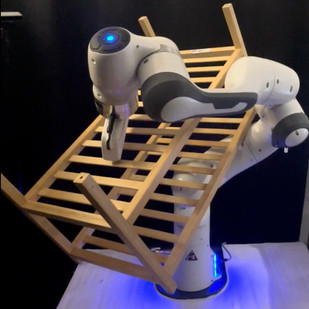} &
  \includegraphics[width=\cw]{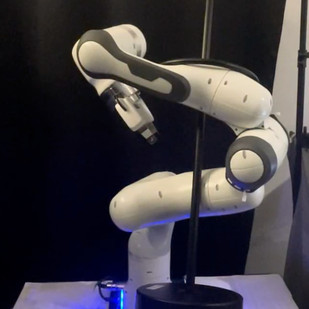} \\[1pt]
  \includegraphics[width=\cw]{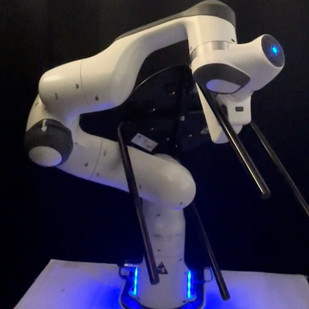} &
  \includegraphics[width=\cw]{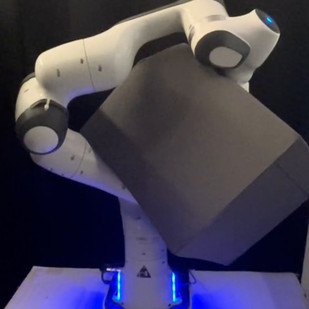} &
  \includegraphics[width=\cw]{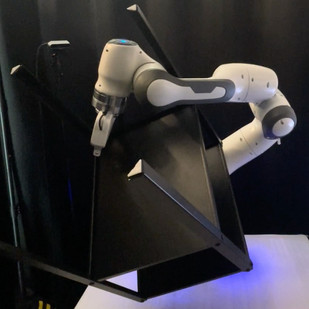} &
  \includegraphics[width=\cw]{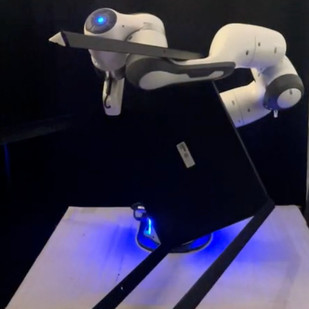} &
  \includegraphics[width=\cw]{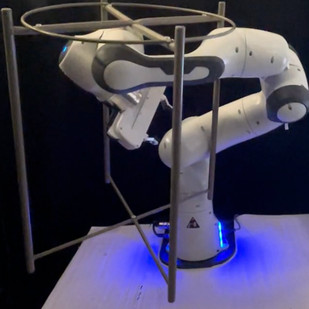} &
  \includegraphics[width=\cw]{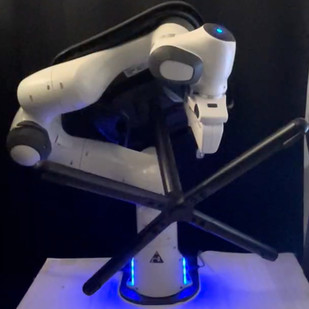} &
  \includegraphics[width=\cw]{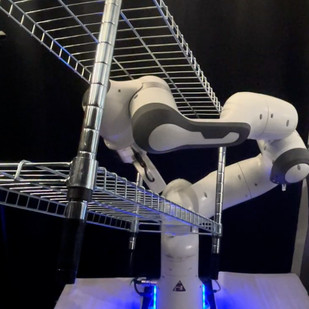} &
  \includegraphics[width=\cw]{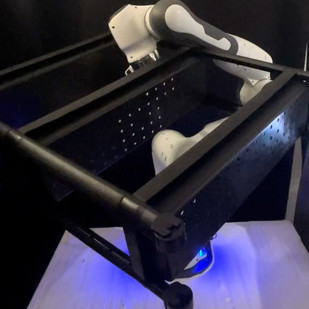} &
  \includegraphics[width=\cw]{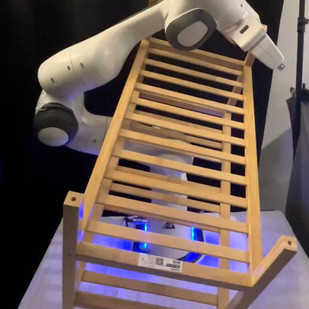} &
  \includegraphics[width=\cw]{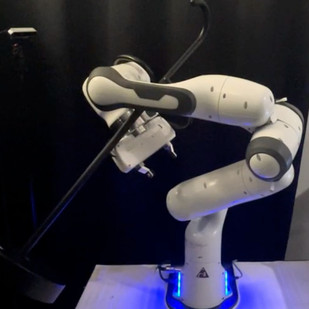} \\
  \scriptsize Object 1 & \scriptsize Object 2 & \scriptsize Object 3 & \scriptsize Object 4 & \scriptsize Object 5\textsuperscript{*} & \scriptsize Object 6 & \scriptsize Object 7 & \scriptsize Object 8 & \scriptsize Object 9 & \scriptsize Object 10 \\
\end{tabular}
\caption{Whole-arm grasps generated by \ours{} on the 10 IKEA
evaluation objects, none of which is in \dataset{} (unlike the dataset
grasps of Fig.~\ref{fig:wag-gallery}). Each column is one object. The
top two rows show two different successful trials in simulation, and
the bottom two show two different successful trials on the physical
robot. \textsuperscript{*}Object~5 appears without its top plate in the
real robot trials. The plate is held to its frame only by two small
adhesive dots and could detach and fall when the arm tilts the object
during the lift, so we removed it for safety. The simulated object
keeps it.}
\label{fig:sim-real-grasps}
\end{figure*}

\begin{figure*}[t]
\centering
\footnotesize
\setlength{\tabcolsep}{1pt}
\setlength{\pwlen}{\dimexpr(0.93\textwidth-12\tabcolsep)/6\relax}%
\begin{tabular}{cccccc}
\includegraphics[width=\pwlen]{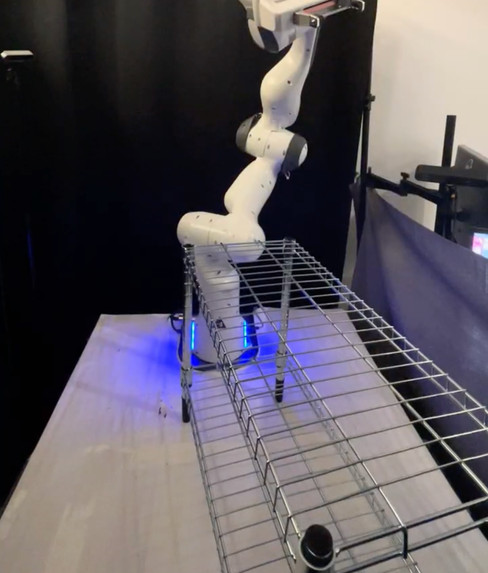} & \includegraphics[width=\pwlen]{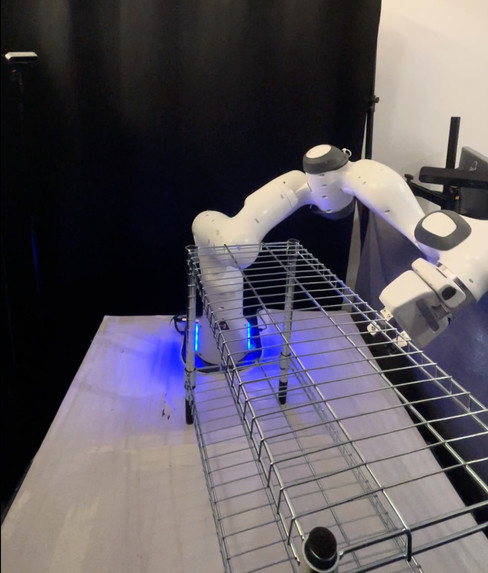} & \includegraphics[width=\pwlen]{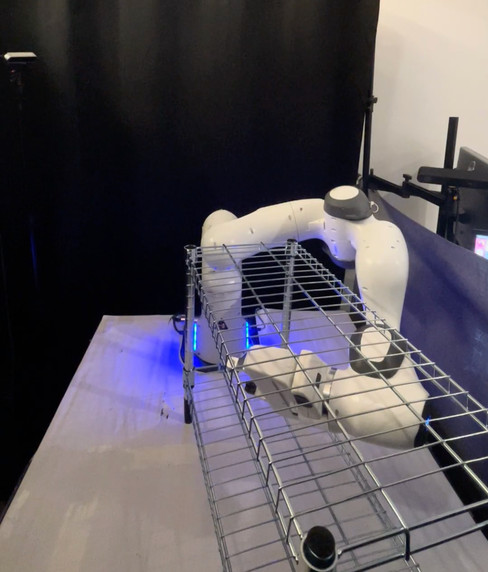} & \includegraphics[width=\pwlen]{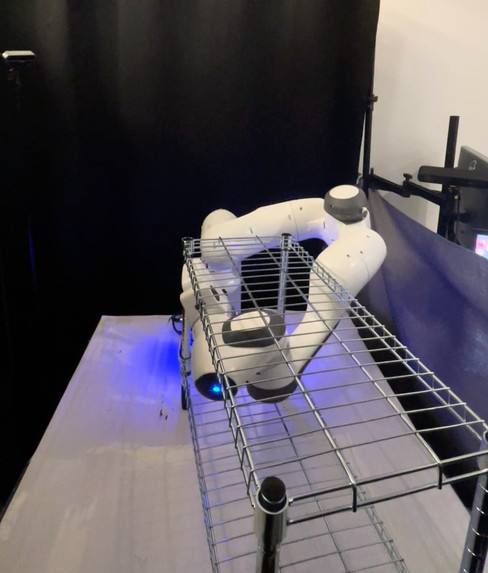} & \includegraphics[width=\pwlen]{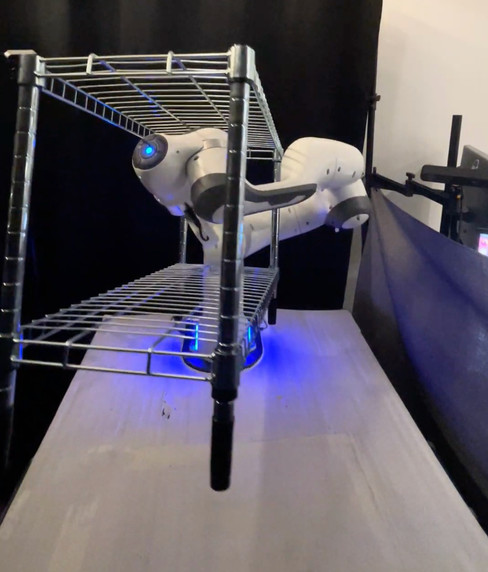} & \includegraphics[width=\pwlen]{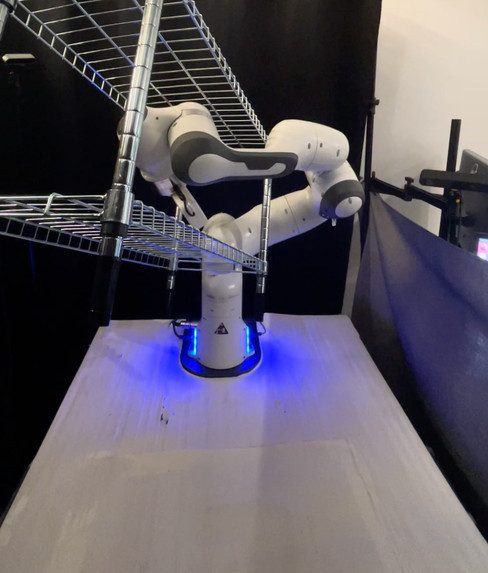} \\[1pt]
\includegraphics[width=\pwlen]{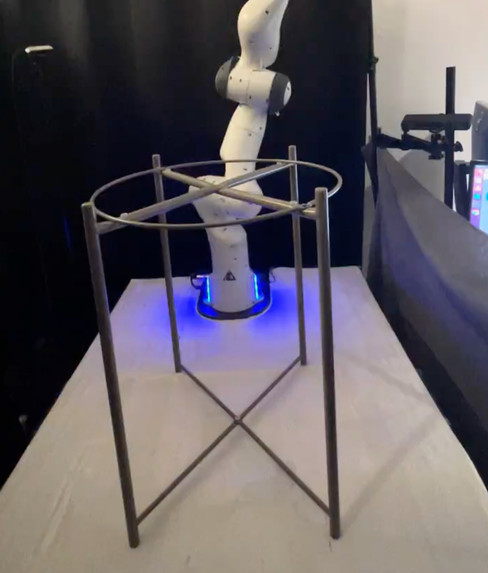} & \includegraphics[width=\pwlen]{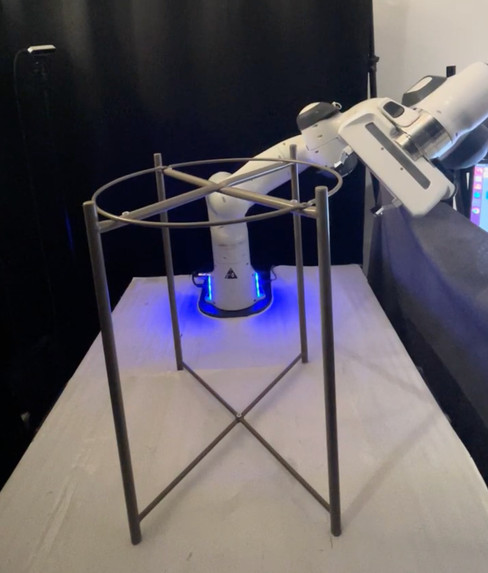} & \includegraphics[width=\pwlen]{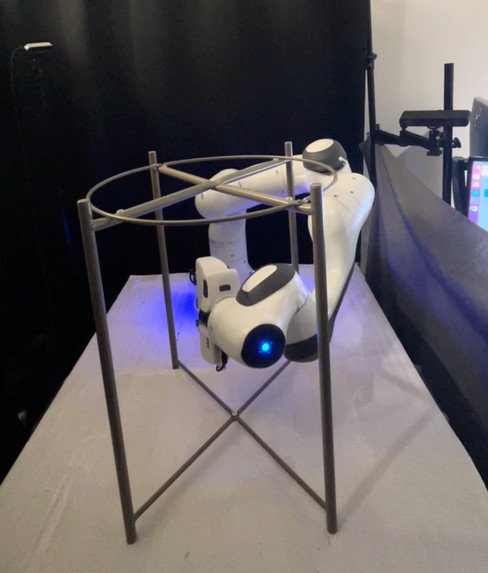} & \includegraphics[width=\pwlen]{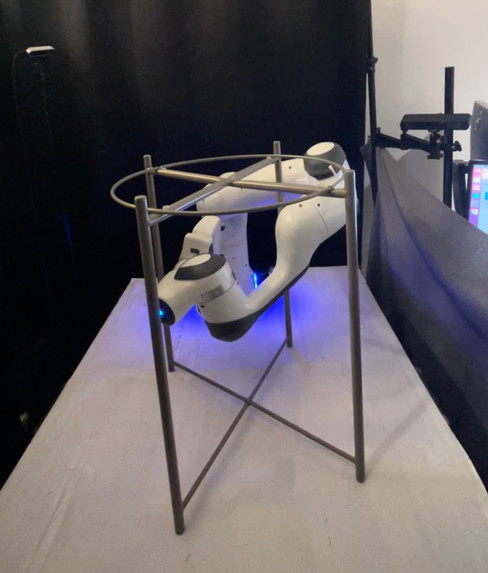} & \includegraphics[width=\pwlen]{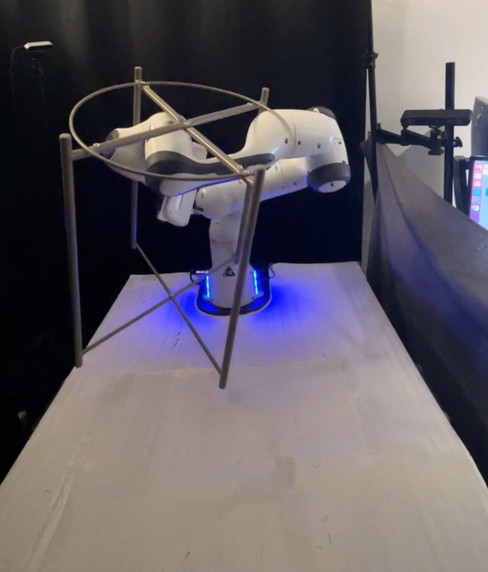} & \includegraphics[width=\pwlen]{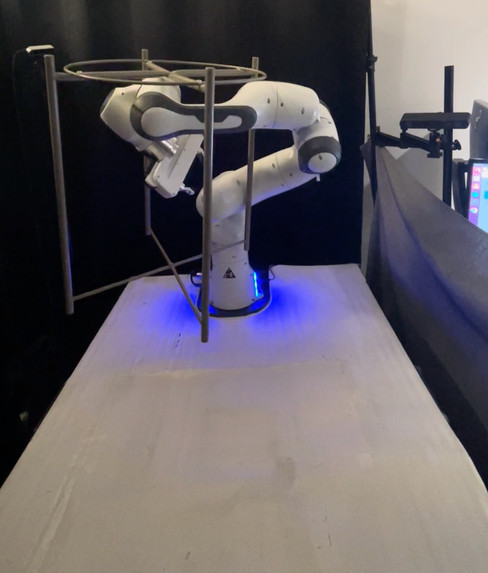} \\[1pt]
\includegraphics[width=\pwlen]{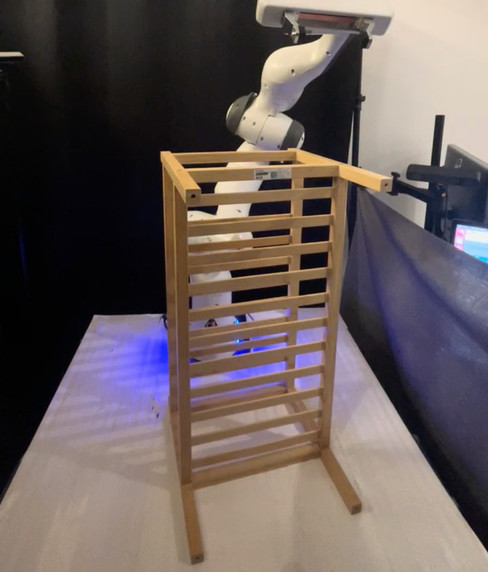} & \includegraphics[width=\pwlen]{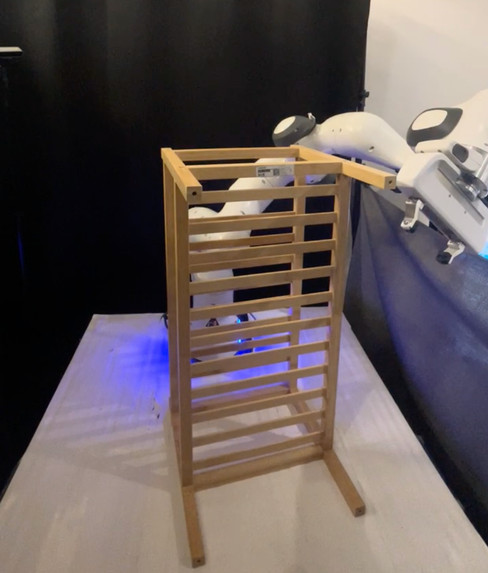} & \includegraphics[width=\pwlen]{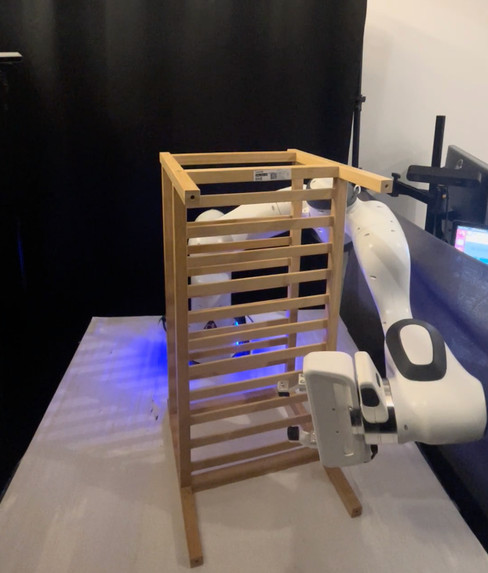} & \includegraphics[width=\pwlen]{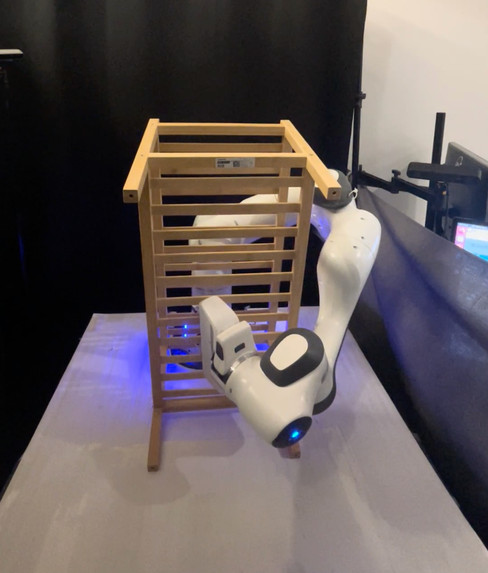} & \includegraphics[width=\pwlen]{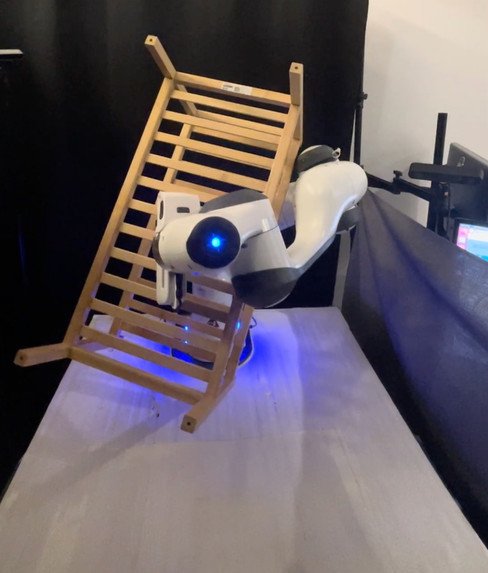} & \includegraphics[width=\pwlen]{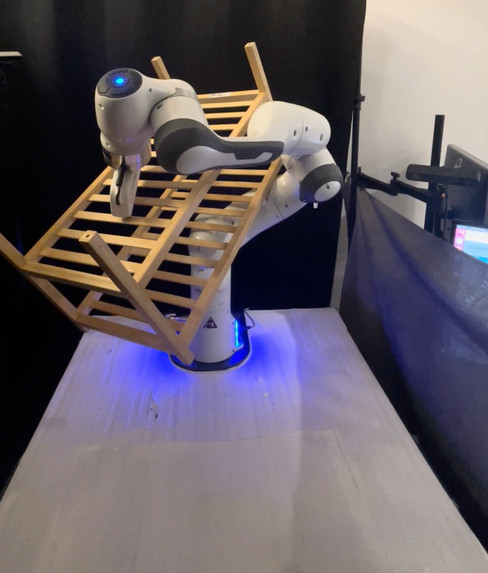} \\[-7pt]
\multicolumn{3}{c}{\phasebrace{\dimexpr 3\pwlen + 4\tabcolsep - 2pt\relax}{Approach}} &
\multicolumn{1}{c}{\phasebrace{\dimexpr \pwlen - 2pt\relax}{Squeeze}} &
\multicolumn{2}{c}{\phasebrace{\dimexpr 2\pwlen + 2\tabcolsep - 2pt\relax}{Lift}} \\
\end{tabular}
\caption{Examples of \ours{} successfully executing a pickup on three objects, from the approach
through the squeeze to the lift.}
\label{fig:pipeline-rollout}
\end{figure*}

We additionally ablate \ours{} by training the same model on
\nocert{}, which skips the stress test of Section~\ref{sec:method-data}.
Its configurations qualify by the nominal lift alone, without the
shake, the external force, and the jittered copies, and we call the
resulting model \ours{} (LiftOnly).
\nocert{} uses the same objects, the same split, the same object
placements, and the same number of configurations per placement as
\dataset{}, so the two differ only in which configurations qualify,
not in how many. Section~\ref{sec:results-q1} reports the effect.

\subsection{Metrics}
\label{sec:setup-metrics}

For every trial, we record \emph{lift success}. In simulation, on all three object
sets, a trial succeeds if the object is raised $10\,\mathrm{cm}$ clear of
the table, and is held there at least $2$\,s, which are the same as criteria for \dataset{} generation. On the physical robot, it succeeds
if the object loses contact with the table and is held there at least $2$\,s.

On the training and held-out object sets we additionally report \emph{success
after perturbation}. The lifted object is subjected to the same shake
and external force used to stress-test \dataset{} in
Section~\ref{sec:method-data}, and the trial counts as a success only
if it is still a \emph{lift success} afterwards. This metric reflects how
firmly the grasp holds the object, not just whether the lift happened.
Some wraps lift the object while leaving it loose inside the arm. They
pass \emph{lift success} but drop the object once it is shaken or
pushed, and are counted as failures here.

We run 5 trials per object per method on the
training object and held-out object sets and 10 trials per object on the IKEA object set,
and report the success rate. On the training and held-out object sets we report its mean and
standard deviation over five evaluation seeds, which change the sampled
point clouds, the model's samples, and the planner's seeds.

\section{Results}
\label{sec:results}

We now present the simulation and real-world results to answer the two questions posed in Section~\ref{sec:setup}.

\subsection{Simulation Performance}
\label{sec:results-q1}

Table~\ref{tab:sim} reports \ours{}, the two baselines of
Section~\ref{sec:setup-baselines}, and \ours{} (LiftOnly) on the training and held-out object sets. Bold marks the best
method per column. \ours{} is the best method on both sets and both
metrics, ahead of nearest-neighbor retrieval by 11.9 and 10.9 points on
the held-out objects. The middle-height wrap is itself a reasonable
strategy, lifting the object in more than half of the trials on both
sets. \ours{} raises lift success over it by 30.8 and 15.6 points on the
training and held-out objects. This gain comes from better
configurations. The wrap closes around the same mid-height
cross-section of every object, whereas \ours{} exploits the geometry of
the object's negative space, such as the openings between legs, rails,
and shelves, and places the arm where it holds the object more
securely. Its lifts are also more robust, with 96\% and 91\% of them
surviving the perturbation on the training and held-out objects,
against 85\% and 84\% for the middle-height wrap.

Training on \dataset{} rather than \nocert{} improves lift success
by 3.9 and 2.5 points across 5 seeds on the training and held-out objects and success
after perturbation by 5.1 and 2.2 points. Since the two datasets are the same size, the stress test helps by
replacing unreliable configurations rather than adding more, and the
gain carries over to held-out objects.

\subsection{Real-World Performance}
\label{sec:results-q2}

We run 10 trials per object on the physical robot for the 10 objects
of Section~\ref{sec:setup-real}. Task difficulty depends largely on
the object and on its placement, both position and orientation, so we
recreate each hardware trial in simulation with the same object at the
same placement, recovering its pose from the point cloud captured for
that trial by iterative closest point (ICP) registration of the object
mesh. This yields 100 paired sim-to-real trials. Table~\ref{tab:q3} reports the results. On hardware,
\ours{} successfully lifts the object in 70 of the 100 trials, with
per-object success rates ranging from 3/10 to 10/10.
Figure~\ref{fig:sim-real-grasps} shows four representative grasps
for each object, two in simulation and two on the physical robot.
Figure~\ref{fig:pipeline-rollout} shows three real-world pickup
sequences, from approach and squeeze through lift.

We then measure the two sides of the transfer separately. The middle row
of Table~\ref{tab:q3} replays the trajectory the hardware executed. The object is never perceived and the policy is
never queried, so this row isolates everything in the transfer except
perception. It lifts 76 of the 100 placements against the robot's 70.
The bottom row runs the full pipeline and lifts 79. Here, the policy is
conditioned on the object's own mesh surface, a complete and exact
shape, placed by ICP where the real object stood. Since the simulated object
sits at the same pose, the row reports what the pipeline reaches
without perception error. In simulation on the same placements,
averaged over five evaluation seeds, the middle-height wrap lifts 67.8\%
and NN retrieval 45.0\% which performs worse than our method.

The three rows give similar totals, but the per-object counts still
differ by as much as 5 of 10 trials on a single object. Object~3 and
Object~6 are heavy with smooth surfaces and slip out of the arm on
hardware, so the robot scores below both simulated rows. Other objects exhibit deformation or articulated motion on hardware
that is not captured by the rigid-body simulation. Object~2 is a
fabric box that deforms around the arm, while the legs
of Object~1, Object~5, and Object~7 rotate about their bolted joints. The
cameras capture its thin frame only in part, so the planner checks
collisions against an incomplete cloud, and it lifts 7 of 10 on the
robot and 5 of 10 in replay against 9 of 10 from the exact mesh.

\section{Limitations}
\label{sec:limitations}

\ours{} has two main limitations: a sim-to-real gap in both perception
and dynamics, and a single hand-designed squeeze-and-lift motion shared
by all objects.

\paragraph{Sim-to-Real Gap} (1) Perception gap. The point cloud goes to
both the grasp model and the planner, so it affects the
grasp configuration and the collision check at once, and because depth
error grows with distance and our objects are large, thin parts come out
incomplete or missing. (2) Dynamics gap. Every object in simulation also gets the same mass and friction
coefficient instead of its real values, so a grasp that holds in
simulation may still let a heavier or smoother real object slip out
of the arm. Furthermore, every object in sim
is rigid, so nothing deforms around the arm and no joint turns under it.

\paragraph{Hand-Designed Squeeze and Lift} Closing the grasp and lifting objects are
easy for a parallel-jaw gripper, a suction gripper, or a dexterous hand, but
not for an arm around a large object, where the squeeze and lift move several
joints at once and the object rotates inside the wrap as it rises. We hardcode the motion as a single wrap by its
shoulder, elbow, and wrist with no regrasp and no adaptation to the object, and
fixed it before generating \dataset{}. A configuration enters \dataset{}
only if it holds the object under that motion, so a model trained on
\dataset{} inherits the same restriction. Learning the motion along with
the configuration would
remove it.

\section{Conclusion}
\label{sec:conclusion}

We introduce \dataset{}, to the best of our knowledge the first
large-scale whole-arm grasp dataset for single-arm pickup of large
objects. \dataset{} contains approximately 1.7 million grasp
configurations across 793 objects and is constructed using a scalable
generation procedure with a perturbation-based stress test, which
retains only configurations that hold the object under shaking and
external forces. Building on this dataset, we develop \ours{}, which
accomplishes the full pickup task by learning whole-arm grasp
configurations from an object's point cloud and the arm's initial
configuration, reaching them with collision-free motion planning, and
executing the same squeeze and lift primitives used to build
\dataset{}. \ours{} outperforms heuristic and retrieval-based
baselines in simulation and successfully lifts objects in 70 of 100
physical-robot trials across 10 objects not included in \dataset{}.
We hope \dataset{} and \ours{} will encourage further work on whole-arm
manipulation of large everyday objects.

\section*{Acknowledgment}

We are grateful to everyone in the RESL Lab at USC, especially
Guangyao Shi, Jeremy Morgan, I-Chun Arthur Liu, Jason Chen, and Ruohai Ge, as well as Yiyang Ling, for their valuable
discussions and feedback on problem formulation, experiments, and
writing. We acknowledge generous support from Samsung Research America. Daniel Seita also acknowledges support from NSF grant \#2447397.

\bibliographystyle{IEEEtran}
\bibliography{references}

\end{document}